\documentclass[runningheads]{llncs}

\usepackage{eccv}

\usepackage{eccvabbrv}

\usepackage{graphicx}
\usepackage{booktabs}

\usepackage[accsupp]{axessibility}  

\usepackage{hyperref}

\usepackage{orcidlink}

\usepackage{booktabs}
\usepackage{multirow}
\usepackage[table]{xcolor}
\usepackage{float}  
\usepackage{graphicx}
\begin{document}

\title{DTI: Dynamic Trajectory Initialization for Generative Face Video Super-Resolution} 

\titlerunning{DTI}

\author{Yingwei Tang\inst{1}\orcidlink{0009-0005-0526-4747} \and
Chen Yan\inst{1}\orcidlink{0009-0003-1580-1708} \and
Wendi Liu\inst{1}\orcidlink{0009-0004-7725-2908} \and
Qiang Hu\inst{1}\orcidlink{0000-0003-4645-9776}\thanks{Corresponding author.} \and
Xiaoyun Zhang\inst{1}\orcidlink{0000-0001-7680-4062}}

\authorrunning{Y. Tang et al.}

\institute{Cooperative Medianet Innovation Center, Shanghai Jiao Tong University, Shanghai, China \\
\email{\{xcn\_tyw, yc.super, lwd624, qiang.hu, xiaoyun.zhang\}@sjtu.edu.cn}
}

\maketitle
\begin{figure}[h!] 
    \centering
    \includegraphics[width=\textwidth]{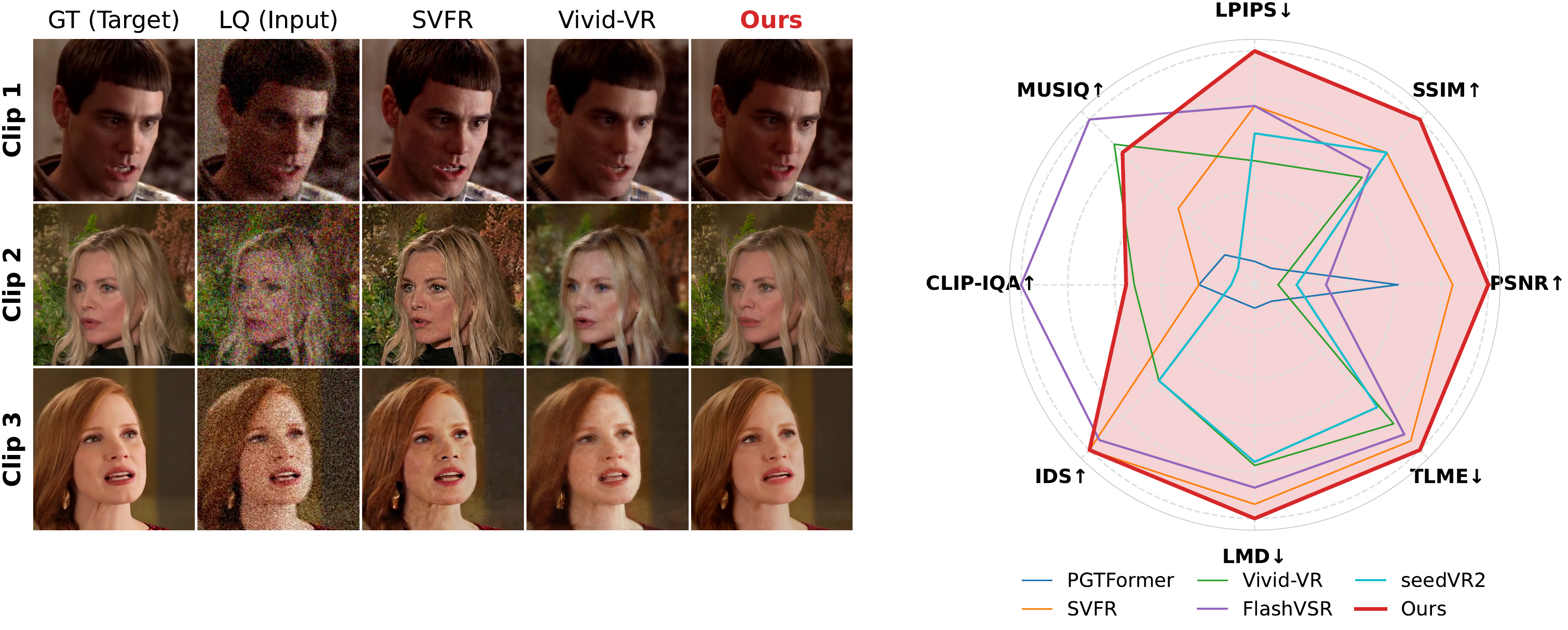}
    \caption{
        \textbf{Left:} Qualitative comparison of video restoration results. 
        \textbf{Right:} comparison using normalized metrics (max-participant-based normalization). Our method achieves the best balance between fidelity and perceptual quality.
        }
    \label{fig:teaser}
\end{figure}

\begin{abstract}

 As the most perceptually powerful Face Video Super-Reso-
 lution (FVSR) method, existing works in Generative FVSR (GFVSR) mainly exploit the generative prior of pretrained diffusion models. However, viewed as full generation, they suffer from fixed sampling and expensive inference costs if without large-scale auxiliary training. Furthermore, an excessive pursuit of generic perceptual metrics often results in low fidelity. To address these issues, we present Dynamic Trajectory Initialization (DTI) paradigm for GFVSR, which reformulates GFVSR as an input-driven directional restoration. With a novel enhancement-and-injection conditioning mechanism for pretrained DiT backbone, fidelity of our model has been significantly improved without compromising perceptual quality. To dynamically set the starting sampling point, we propose a Discriminative Guide (DG) trained via objective Signal-to-Noise Ratio (SNR) alignment. With only minor model adaptation and fine-tuning, our method achieves a SOTA overall performance across diverse metrics and benchmarks. An analysis of relationship between actual comprehensive quality and common metrics is also conducted, which demonstrates the perception-distortion trade-off and that the LPIPS is the most convincing metric in our case. Code: \url{https://github.com/MediaX-SJTU/DTI}
  
\end{abstract}

\section{Introduction}
\label{sec:intro}
With its wide-ranging applications such as restoring speeches in news media, enhancing historical footage, and reconstructing forensic evidence, \textbf{Face Video Super-Resolution} (FVSR) presents a highly challenging ill-posed problem, for it must recover high-quality spatial-temporal information close to the ground truth from unknown real-world degradations. Formally, FVSR focus on reconstructing high-quality face videos (HQ) from degraded low-quality inputs (LQ). In these years, \textbf{Generative Face Video Super-Resolution} (GFVSR) present the best restoration results, but numerous issues remain.

Face Super-Resolution has long been an essentially important field in computer vision. Based on data types, we can classify it into: face image super-resolution (FISR) processing static images and FVSR treating videos with temporal dimension. Early FISR studies attempted to leverage various types of facial prior information, such as geometric priors\cite{CT-FSRNet-2018, Yu_2018_ECCV}, reference priors\cite{Li_2020_ECCV, Li_2020_CVPR} and generative priors\cite{chan2021glean, Yang2021GPEN}. Despite the reconstruction of majority low-frequency information in one-step mapping, they struggle to restore high-quality details due to the irreversibility of information loss, the focus on certain priors also reduces their robustness when applied to more general real-world scenarios. This direction can be referred to as \textbf{discriminative}. With the continuous advancement of large generative models, especially diffusion models, numerous high-performance generative FISR have been developed\cite{9588816, dr2, fang2026one}. The powerful generative prior of diffusion models has significantly improved the visual quality of restored details. In contrast to extensively studied FISR, there are less research attention towards FVSR, yet it is an increasingly important field today as videos become a central component of streaming media and social networks. For FVSR, main works are still conventional discriminative\cite{ChenPSFRGAN, feng2024keep, xu2024beyond}, several works explored the generative approaches\cite{Yang2021GPEN, wang2025svfrunifiedframeworkgeneralized}, but the majority among them conduct multi-frame restoration using existing FISR models or directly adapting general Generative Video Super-Resolution (GVSR) models\cite{xie2022vfhq, chen2023realworld}.

The most powerful GFVSR methods undoubtedly surpass other discriminative approaches, yet it also inherits the shortcomings of diffusion-based Video Super-Resolution (VSR) techniques. Similar to advanced GVSR models\cite{RealisVSR, SeedVR, VividVR, RealisVSR, FlashVSR}, GFVSR models also necessitate a huge inference computation burden, heavy auxiliary training or distillation owing to characteristics of diffusion-based multi-step sampling. Furthermore, the face-unique characteristics of structural clarity and well-defined subject areas have not been sufficiently exploited for deep mining of LQ information. Despite enhanced detail generation capabilities from pretrained models, fidelity has not been assured in the reckless pursuit of perceptual quality. This phenomenon can be formulated as distribution generation trap: emergence of unnatural distortions or repetitive visual artifacts, resulting in a lower consistency with ground truth (GT).

To address the aforementioned issues and to enhance GFVSR, in this work we propose a \textbf{Dynamic Trajectory Initialization} (DTI) paradigm for GFVSR. We argue that the GFVSR process should not be regarded as a purely generative task as in previous studies, but rather as a dynamic restoration process driven by the extent of information loss in the input: preserving existing GT information and repairing damaged information. In other words, this paradigm transforms GFVSR from a simple conditional generation into a directional restoration, which places greater emphasis on fidelity and consistency with facial information, as well as authentic perceptual quality rather than pursuing improvements on generic metrics. 

Given that LQ is lossy information, we innovatively introduce visual feature extractors into the framework, for face video features are highly suitable for their application. By redesigning the conditioning methods of generative approaches along with adaptation of model architecture, we obtain a GFVSR model with largely enhanced fidelity and reduced training overhead (accelerated model convergence) while maintaining the theoretical consistency with flow-matching diffusion. By theoretically deriving new training paradigm of discriminative components through objective signal-to-noise ratio alignment, we realize an \textbf{interpretable and controllable dynamic-starting restoration framework}. An analysis of relationship between metrics and true comprehensive quality is also carried out after the evaluation.

We conclude our main contributions below:
\begin{itemize}
    \item We present Dynamic Trajectory Initialization (DTI) paradigm for GFVSR, reformulating GFVSR from “full generation” to “input-driven restoration”, successfully decoupling it from pure generation tasks.
    
    \item We introduce novel enhancement-and-injection method for model conditioning with extracted visual features and conditions parallelism. We propose a Discriminative Guide (DG) and design its interpretable training by objective supervision, which estimates the reasonable intermediate distribution of diffusion trajectory which LQ should belong to.
    
    \item Through only minor model adaptation and fine-tuning, our approach achieves state-of-the-art performance with significantly improved efficiency and fidelity. By analyzing phenomena observed during evaluation, we also discuss the relationship between metrics and actual quality, demonstrating the optimality of LPIPS in perception-distortion trade-off for this task.
\end{itemize}

\section{Related Work}

\subsection{Face Super-Resolution}

Face Super-Resolution aims to recover ground-truth HQ from their LQ counterparts, where LQ stems from the intractable degradations in the real-world. Formerly, the highly structural nature of human faces allows researchers to incorporate different types of prior information into restoration models or frameworks: geometry priors such as parsing maps or facial landmarks\cite{ChenPSFRGAN, CT-FSRNet-2018}, reference-based priors such as high-quality images\cite{Li_2018_ECCV, Li_2020_CVPR}, and generative priors including codebook priors\cite{gu2020image, pan2020dgp, gu2022vqfr, zhou2022codeformer, xu2024beyond, feng2024keep, Zhang2025TDBFRTD} such as pretrained StyleGAN model\cite{StyleGAN}. Most early methods tend to produce low visual quality and lack of universality. Recently, restoration methods based on advanced deep neural networks or large models have been developed\cite{wang2025svfrunifiedframeworkgeneralized}, some adapt directly general VSR networks to face video datasets\cite{xie2022vfhq}, some extend the image-based models or approaches to video tasks\cite{BasicVSR++, feng2024keep}. 

\subsection{Diffusion-based Video Generation Models}
Diffusion models tends to generate clean data from Gaussian noise by gradually denoising noisy data. Based on score matching or flow matching, probabilistic or deterministic diffusion models represent the most powerful visual generation models available today. From the original UNet-based backbones (DDPM/LDM)\cite{ho2020denoising, rombach2021highresolution} to the current transformer-based backbones (DiT)\cite{Peebles2022DiT}, architectural advancements have enabled scalability, paving the way for large-scale video diffusion generative models. These models can learn high-quality generative patterns for video data, which possesses both spatial and temporal dimensions. In recent years, continuous efforts from the open-source community have steadily narrowed the gap between open-source and closed-source video generation models\cite{yang2024cogvideox, kong2024hunyuanvideo, wan2025}. 

\subsection{Real-World Video Super-Resolution}
Early researches in general VSR primarily focused on predefined idealized degradations, their performance deteriorate significantly when handling unknown de-
gradations in reality\cite{VSRDUF, EDVR, chan2022investigating}. In recent years, the development of diffusion-based large generative models has ultimately enabled a leap in visual quality for high-resolution VSR. Their powerful generative pattern learning abilities allow fine-tuned models to produce acceptable restoration results for complex real-world degradations, but also inherit their shortcomings. Extensive inference cost, insufficient fidelity and unnatural artifacts remain critical challenges after simply standard fine-tuning\cite{SeedVR, VividVR, RealisVSR}. Heavy auxiliary components training and model post-training using large-scale data\cite{wang2025seedvr2, FlashVSR} deviate from the inherent difficulty of restoration problem, offering poor cost-effectiveness.

\section{Reasoning and Design}
\subsection{Preliminaries: Diffusion Transformer as Flow-Matching Diffusion Model}
\label{3.1}
Unlike original probabilistic diffusion models based on stochastic Langevin dynamics\cite{song2021scorebased}, flow-matching (FM) diffusion models simulate the generative process of video latents by deterministically transforms a simple prior noise distribution $p_1$ (e.g., $\mathcal{N}(0, I)$) into the complex data distribution $p_0$ via an Ordinary Differential Equation (ODE)\cite{flowmatching}.The trajectory of data samples is governed by a time-dependent vector field $v_t$, defined by the ODE $dx_t/dt = v_t(x_t)$, where $t \in [0, 1]$. Usually, Conditional Flow Matching objective with an Optimal Transport path\cite{rectifiedflow} are used to enable efficient training, which corresponds to a linear interpolation between the high-quality data $x_0$ and the noise $x_1 \sim \mathcal{N}(0, I)$. Specifically, the intermediate state $x_t$ at time $t$ is constructed as:
\begin{align}
    x_t = (1 - t)x_0 + t x_1.
\end{align}

This linear path implies a constant target velocity field for a given data-noise pair. Differentiating the path with respect to $t$ yields the ground-truth velocity $u_t(x|x_0, x_1) = x_1 - x_0$. Consequently, a neural velocity estimator $v_\theta$ conditioned on time $t$ and other conditions $c$ is trained to regress this target direction. The training objective is to minimize the expected mean squared error between the predicted velocity and the ground-truth vector pointing from data to noise:
\begin{align}
    \mathcal{L}_{FM}(\theta) = \mathbb{E}_{t, x_0, x_1} \left[\omega(t)\| v_\theta(x_t, t, c) - (x_1 - x_0) \|^2 \right],
\end{align}
where $v_\theta$ is predicted velocity, $\omega(t)$ represents a time-dependent weight. During inference, output is sampled by solving the ODE from noise $x_1$ to $t=0$, guided by velocity field $v_\theta$ and conditions $c$ using a numerical solver (e.g., Euler method).

Diffusion Transformer (DiT) \cite{Peebles2022DiT} is a diffusion model design with transformer backbone. Its conditional controllability during generation primarily stems from the attention blocks within the transformer structure. This mechanism enables the sampling entity to compute weighted attention over conditions, the condition information is then injected as residual increments.


\subsection{Conditional Generation as Restoration}
\label{3.2}
\begin{figure}[tb]
  \centering
  \begin{subfigure}{0.25\linewidth}
    \includegraphics[width=\linewidth]{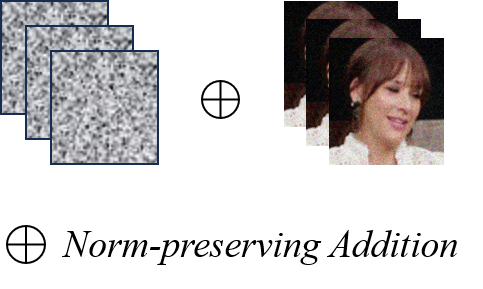}
    \caption{Addition}
    \label{fig1::a}
  \end{subfigure}
  \hfill
  \begin{subfigure}{0.25\linewidth}
      \includegraphics[width=\linewidth]{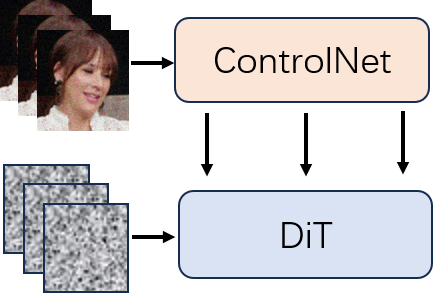}
      \caption{ControlNet}
      \label{fig1::b}
  \end{subfigure}
  \hfill
   \begin{subfigure}{0.24\linewidth}
      \includegraphics[width=\linewidth]{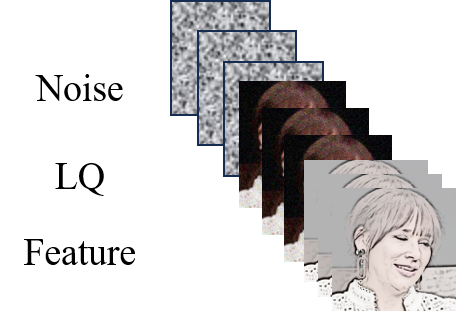}
      \caption{Our Conditioning}
      \label{fig1::c}
  \end{subfigure}
  \hfill
  \begin{subfigure}{0.22\linewidth}
      \includegraphics[width=\linewidth]{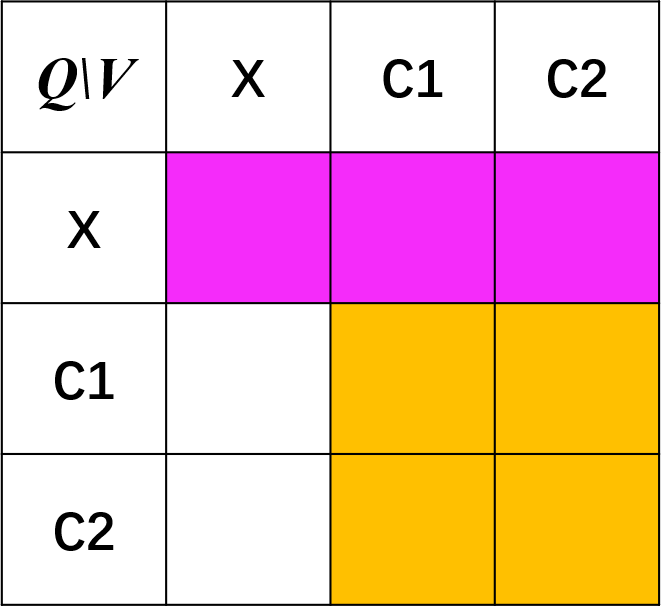}
      \caption{Attention Matrix}
      \label{fig1::d}
  \end{subfigure}
  \caption{(a)(b) some conventional conditioning methods; (c) our novel condition injection method; (d) represents the real calculated attention parts in adapted DiT, \textit{\color{orange}orange} ones are only calculated once in a forward.}
  \label{fig1}
\end{figure}

To transform a DiT-based diffusion generation model into a VSR model, past studies have already some practices: by modifying the starting point of denoising process, it becomes a norm-preserving linear combination (\cref{fig1::a}) of pure Gaussian noise and LQ\cite{FlashVSR}; by adopting ControlNet architecture (\cref{fig1::b}), learnable control networks embed LQ and inject information into diffusion backbone as an additional factor to timestep condition\cite{RealisVSR}; adopting both noisy LQ as input and ControlNet-style injection mechanism\cite{VividVR}; concatenate LQ and noise then passing it into MLP layers after flattening\cite{SeedVR, wang2025seedvr2}. They each have their own shortcomings: under standard sampling, linear combination fails to align with diffusion model due to input shifting; ControlNet requires heavy auxiliary components and training; concatenation-flattening-MLP only performs channel-level information exchange, resulting in relatively low quality. Besides, all previous works directly employ LQ without considering the possibility of extracting information from it to enhance the condition.

The primary consideration is that LQ, as our sole information source, is inherently lossy data. Much useful information lies buried beneath degradation such as blurring or noise, making it insufficient as a direct condition. Naturally, to enhance this condition, \textbf{visual feature extraction models} occur to our mind: human faces are highly suitable for application of such networks, for they possess strong visual structural characteristics such as distinct main areas and clear subjects. Theoretically, as extractors embed patches, extracted features not only are enhanced visual information, but also can be seen partially purified.

Secondly, to efficiently and succinctly inject conditions while maintaining alignment with Diffusion theory, we propose connecting pure noise (the sampling subject) end-to-head with condition tokens, treating them as \textbf{one input sequence} to DiT (\cref{fig1::c}), and achieving conditional generation through attention calculations. Moreover, based on the fact that conditions don't need to be influenced by noise, bidirectional attention inside conditions and unidirectional attention between noise and conditions can realize sufficient conditioning and reduce the computational burden (\cref{fig1::d}), for the complexity of bidirectional attention increases quadratically with the length of sequence.
\subsection{Objectively Dynamic Initialization by SNR Alignment}
\begin{figure}[tb]
  \centering
  \begin{subfigure}{0.4\linewidth}
    \includegraphics[width=\linewidth]{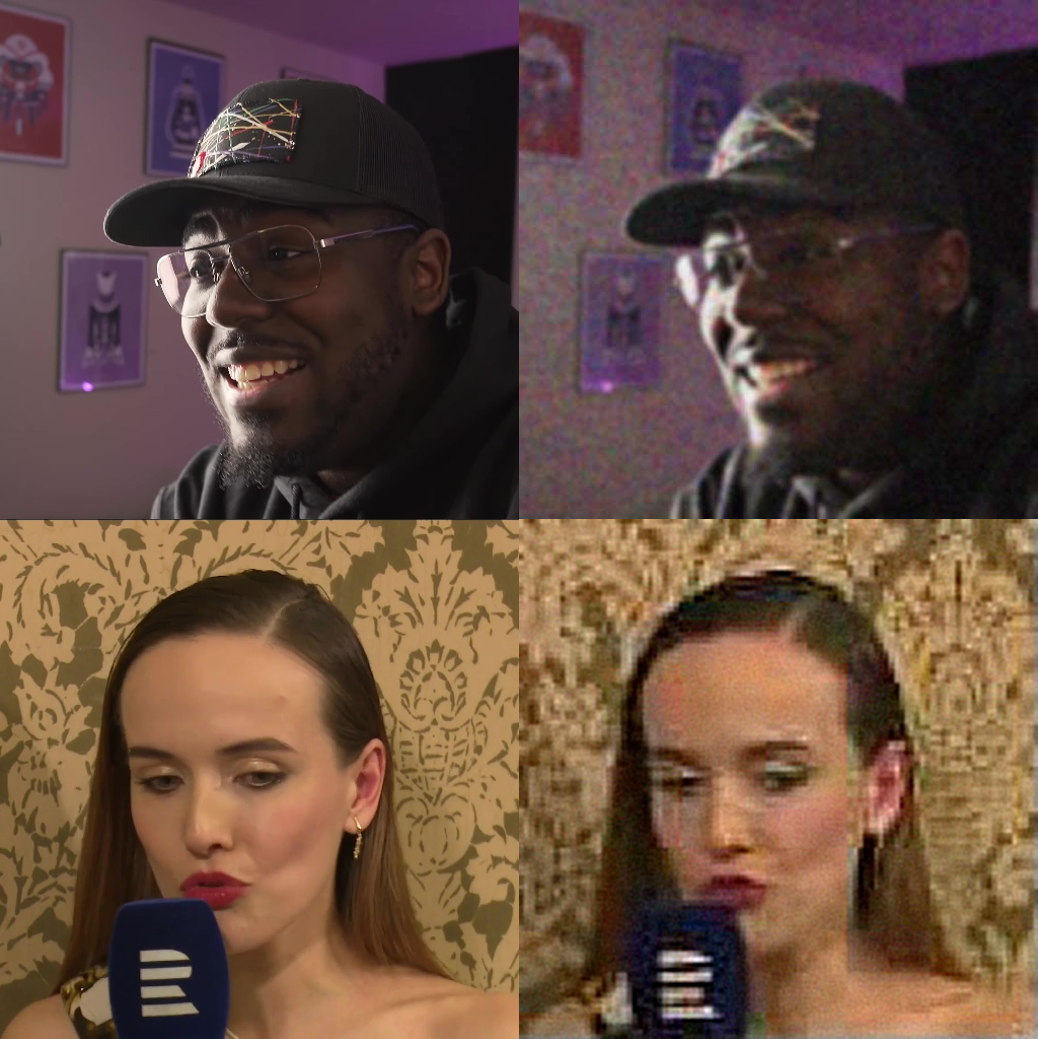}
    \caption{Low-frequency Info Consistency}
    \label{fig2::a}
  \end{subfigure}
  \hfill
  \begin{subfigure}{0.58\linewidth}
      \includegraphics[width=\linewidth]{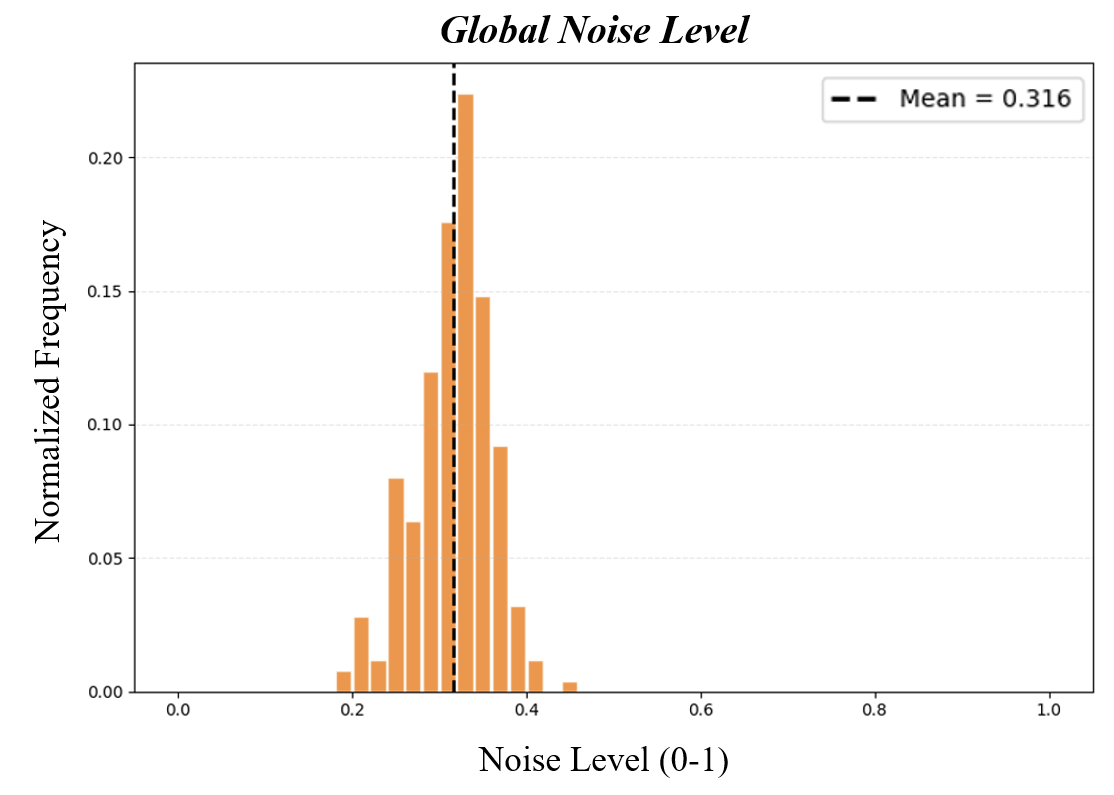}
      \caption{SNR-aligned Noise Level}
      \label{fig2::b}
  \end{subfigure}
  \caption{(a) represents that LQ preserves perceptually more low-frequency information than high-frequency; (b) visualizes the SNR-aligned timesteps corresponding to real-world degradations on a test subset.}
  \label{fig2}
\end{figure}
As the examples show (\cref{fig2::a}), previous related work\cite{dr2} has observed that LQ retains much low-frequency information (overall visual structure, regional colors) from GT, with the primary information loss occurring in high-frequency domain (texture features, edge sharpness). This observation supports that mapping neural networks (NN) in discriminative methods suffice to an efficient coarse-grained refinement in low-frequency domain with acceptable quality, while struggling to restore high-frequency details. This insight inspires us to employ a discriminative approach to guide generative restoration, ensuring proximity to low-frequency ground-truth information and avoiding unnecessary generative sampling.

Inspired by ideas of related studies\cite{wu2025omgsrneedmidtimestepguidance} in vision field, the \textbf{signal-to-noise ratio} (SNR) is used in this work to objectively estimate the relationship between a diffusion timestep which represents the injection proportion of Gaussian noise and the global degradation level of real-world randomly degraded samples. Here, we extend theoretically the proposal:

As mentioned in \cref{3.1}, noisy latent $z_t$ at timestep $t$ in diffusion process is a linear combination of Gaussian noise $\epsilon$ and the ground-truth HQ latent $z_H$. Similarly, LQ are obtained by applying real-world degradation to HQ, where $z_L$ can be regarded as combination of residual difference $z_L-z_H$ and $z_H$. Their SNR can be expressed as:
\begin{align}
    SNR(z_t) = \frac{(1-t)^2 \mathbb{E}\left[z_{H}^{2}\right]}{t^2 \mathbb{E}\left[\epsilon^{2}\right]},  
    \label{3}
    \\
    SNR(z_L) = \frac{\mathbb{E}\left[z_{H}^{2}\right]}{\mathbb{E}\left[(z_L - z_H)^{2}\right]},
    \label{4}
\end{align}
where $\epsilon \sim \mathcal{N}(0, I)$ and $\mathbb{E}\left[\epsilon^{2}\right] = 1$. Defining $RMSE(x,y)= \sqrt{\mathbb{E}\left[(x-y)^{2}\right]}$ as square root of mean square error, by linking \cref{3} and \cref{4}, we can derive the following equation:
\begin{align}
    t = \frac{RMSE(z_L, z_H)}{1+RMSE(z_L, z_H)}.
    \label{5}
\end{align}

In \cref{fig2::b}, the noise level distribution (equivalent to timestep in diffusion process) calculated using \cref{5} effectively demonstrates the preservation of low-frequency information in LQ, further validating the significance of discriminative component design. The global noise level of real-world degradations are SNR-equivalently to the 20\%-45\% of Gaussian diffusion process.

\section{Methodology}
\subsection{Conditioning and Architecture of GFVSR Model}
\begin{figure}[tb]
  \centering
  \includegraphics[width=\linewidth]{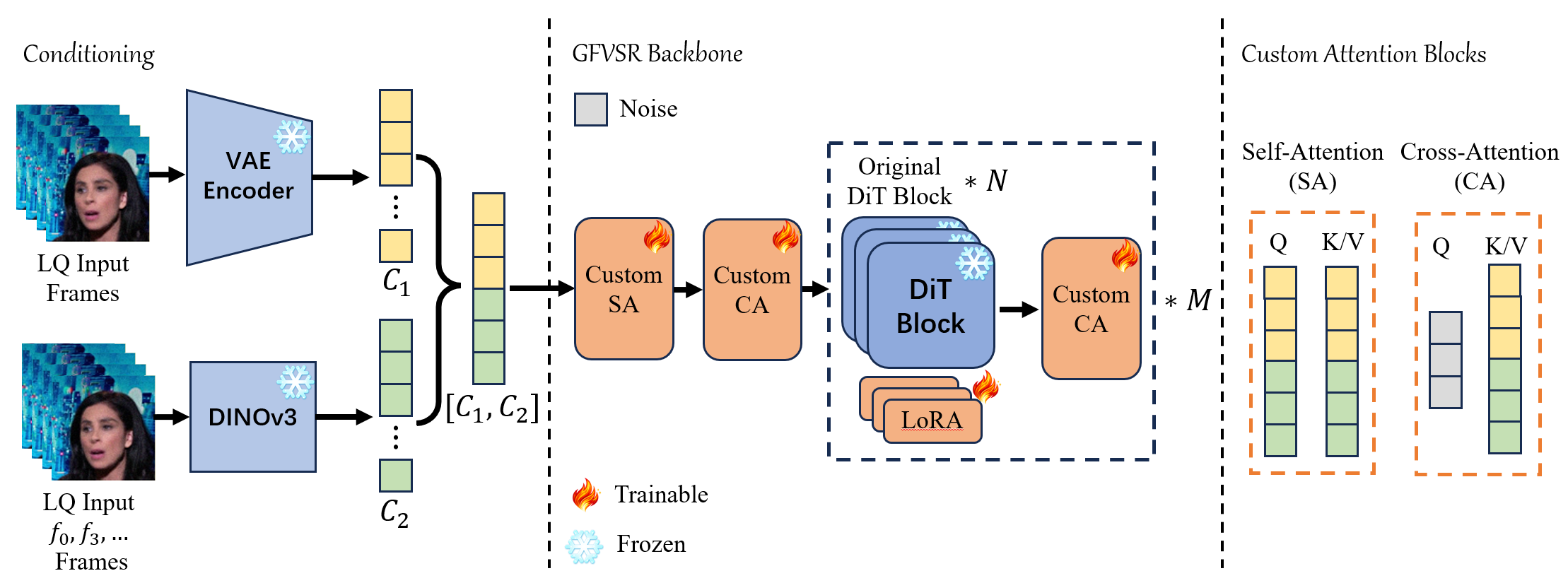}
  \caption{Model Structure \& Pipeline}
  \label{fig3}
\end{figure}

Human faces exhibit a clearly defined subject, an overall structured composition and pronounced edges.
Given the need to extract as much useful visual information as possible from LQ, \textbf{fine-grained visual feature extractors} are our preferred choice. Compared to semantic-aligned contrastive learning models like CLIP\cite{radford2021learningtransferablevisualmodels}, the DINO family of self-supervised pure vision models clearly better fits our requirements\cite{oquab2023dinov2}. Since the VAE encoder also downsamples the input along the temporal dimension, to naturally align with the sequence shape, we apply the feature extractor to the intermediate frames at the downsampling intervals, serving as a pillar regularization for the overall trend (\cref{fig3}).

In addition to \cref{3.2}, to further minimize modifications to the backbone, we let subsequence (LQ,Extracted-Feature) do only \textit{one} self-attention computation at the beginning of a model forward. As shown in \cref{fig3}, formally, input sequence is (X, C1, C2), where X represents pure noise tokens, C1 denotes LQ tokens, and C2 signifies extracted feature tokens. One block is added at top of DiT to realize self-attention of (C1, C2) and first cross-attention of X to (C1, C2). A cross-attention block of X to (C1, C2) is also added after every several original DiT blocks, which is equivalent to a unidirectional attention block. It should be noted here that testing confirms our method doesn't make observable change in model's wall-clock time per forward.

\subsection{Dynamic Initialization}
Our conditioning method ensures the consistency with Diffusion theory, yet the sampling process is expensive. Previous studies on diffusion models indicate that during denoising, different timesteps primarily target distinct domains in information: high-noise steps construct low-frequency information first, while low-noise steps focus on generating high-frequency details\cite{ho2020denoising, rombach2021highresolution, sd3}. As LQ already closely matches GT in low-frequency domain, it should not require a complete generation process starting from pure noise. But it can't be treated directly as an intermediate state of noisy latent neither, as LQ is in untractable real-world degradation distribution, not the specific parameterized distribution (Gaussian distribution) learned by diffusion model. So, if we can transfer LQ to a reasonable nearest point in the probability flow field, the generation process can be reformulated to a restoration process with dynamic starting point. 

\begin{figure}[tb]
  \centering
  \includegraphics[width=\linewidth]{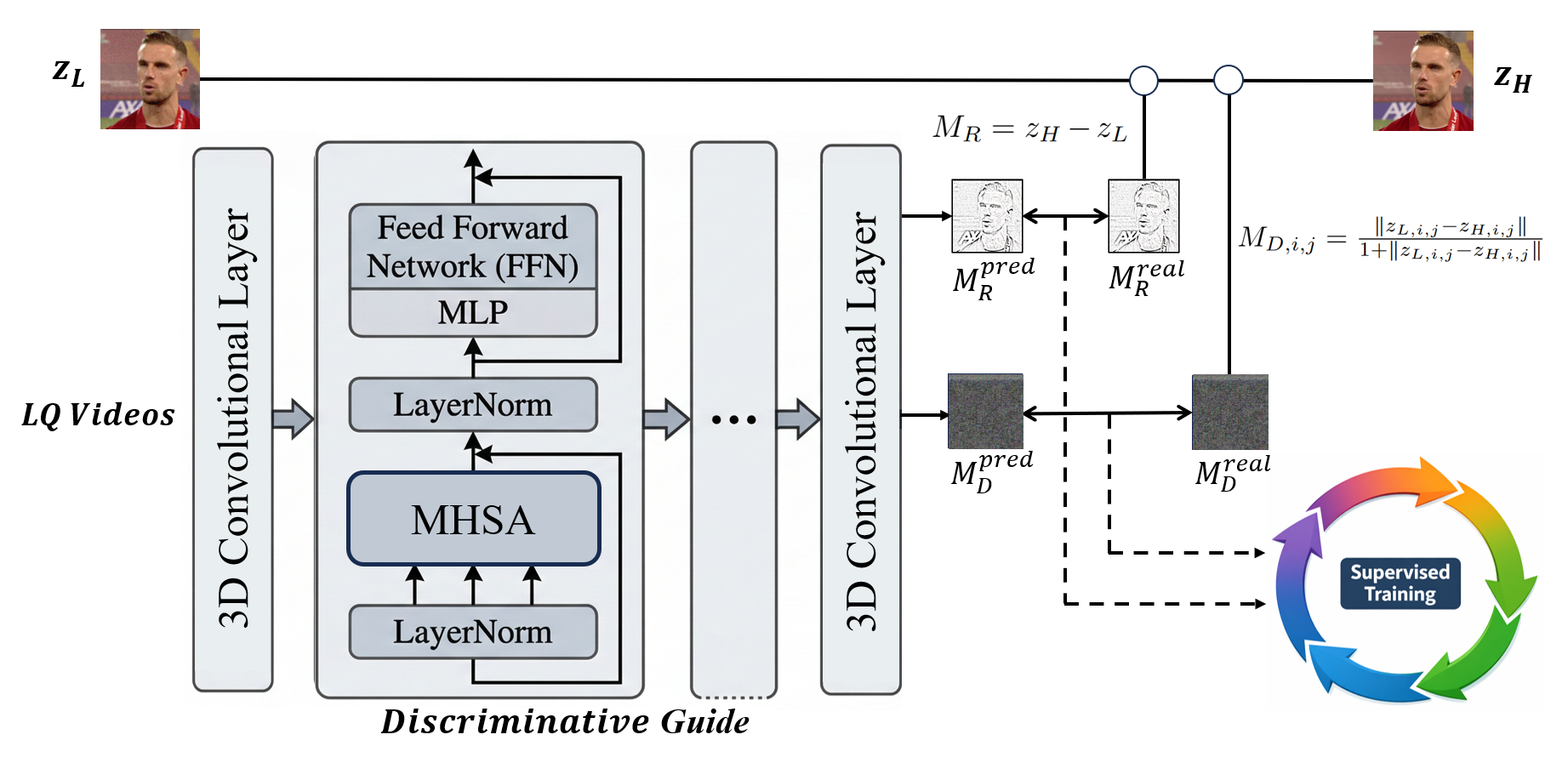}
  \caption{Supervised Training for Discriminative Guide}
  \label{fig4}
\end{figure}

Reasoning in \cref{3.2} provides us with the validation and strategy for a supervised training of \textbf{discriminative guide} (DG). (\cref{fig4}) DG is a lightweight mapping NN to estimate two latents with both the same shape as target latent: one for global noise-level, another for efficient low-frequency refinement. The reason of full-dimension design for first latent is to alleviate the error of global prediction. The first latent is trained to approximate the matrix $M_D$ where $M_{D,i,j} = \frac{\|z_{L,i,j}-z_{H,i,j}\|}{1+\|z_{L,i,j}-z_{H,i,j}\|}$, which represents the local information-loss level; the second to approximate the element-wise residual between HQ and LQ $M_R = z_H - z_L$. As the lightweight DG has limited capability, the predicted $M_R$ is with high error which can only serves as a low-frequency coarse-grained refinement. This objective supervision makes our DG training more interpretable, which is never concerned in any previous work for restoration guidance. 

The dynamic initialization is realized in the following framework: 

The predicted starting timestep is calculated by $M_D$: 
\begin{align}
    t_{pred} = \frac{RMSE(z_L, z_H)}{1+RMSE(z_L, z_H)} = \frac{\sqrt{\mathbb{E}\left[(\frac{M_D}{1-M_D})^2\right]}}{1+\sqrt{\mathbb{E}\left[(\frac{M_D}{1-M_D})^2\right]}}.
\end{align}
The noise proportion in the $z_{start}(t_{start})$ determines a trade-off: relatively, more sampling steps with noisier starting point yield higher perceptual quality but lower fidelity, and vice versa. Therefore, by setting 
\begin{align}
    t_{start}=(1-\lambda) t_{pred} + \lambda t_{max},
\end{align}
where $t_{max}=1, \lambda \in [0,1]$, we obtain a hyperparameter that can flexibly adjust the trade-off: 1 to perceptual quality, 0 to fidelity. We calculate the starting latent before DiT sampling as:
\begin{align}
z_{start} = (1-t_{start}) z_{anchor} + t_{start}\epsilon, 
\end{align}
where $z_{anchor}=z_L+M_R$ and $\epsilon \sim \mathcal{N}(0, I)$.

In this manner, on one hand we employ discriminative prediction to efficiently influence the repair starting point of generative sampling, bringing it closer to HQ while preserving more low-frequency information that does not require much refinement; on the other hand, by initiating from a reasonable timestep (SNR aligned), we leverage DiT's robust generative prior and multi-step denoising characteristics to mitigate the adverse effects of discriminative prediction errors on the final outcome.

\section{Experiments}
\subsection{Datasets and Metrics}
\paragraph{\textbf{Training Data}}
All our training are implemented on the train split of \textbf{VFHQ} dataset\cite{xie2022vfhq}, which includes 16k high quality face clips as GT. During training, the LQ video data are synthesized by the standard Real-ESRGAN degradation pipeline\cite{chan2022investigating, wang2021realesrgan} with lightweight random temporal corrupt: stochastic multi-frame pooling and frame replication to simulate frame drops and temporal stuttering, enabling a more realistic approximation of real-world video degradation. 

\paragraph{\textbf{Test Data}}
Evaluation of our GFVSR model is conducted on three benchmarks: the official \textbf{VFHQ} test dataset \cite{xie2022vfhq}, \textbf{CelebV-HQ} \cite{zhu2022celebvhq} for fine texture recovery and \textbf{VoxCeleb2} \cite{chung2018voxceleb2} for severe in-the-wild degradations. As our DG is only trained from scratch on VFHQ dataset with a small amount of iterations (15k), we conducted its evaluation (ours w/ DG) on corresponding test dataset. 

\paragraph{\textbf{Metrics}}
For datasets with GT, we adopt widely used \textbf{PSNR}, \textbf{SSIM}, and \textbf{LPIPS}\cite{zhang2018perceptual} as general full-reference metrics to evaluate the fidelity; \textbf{IDS}\cite{8953658} quantifying the biometric consistency of the reconstructed faces,\textbf{LMD}\cite{Wang_2019_ICCV} evaluating the alignment accuracy of facial landmarks and \textbf{TLME}\cite{xu2024beyond} measuring the consistency of facial dynamics are evaluated as facial full-reference metrics. We adopt also generic no-reference metrics on all test datasets: \textbf{MUSIQ}\cite{ke2021musiq} for multi-scale aesthetic perception, and \textbf{CLIP-IQA}\cite{wang2022exploring} for high-level semantic quality.


\subsection{Implementation}
The backbone of our GFVSR model is the official Wan2.1-1.3B T2V DiT model\cite{wan2025}. The selected visual feature extractor is DINOv3\cite{simeoni2025dinov3}, embedding the first and every third among four frames in the rest, aligned with temporal downsampling of Wan VAE encoder. All experiments are implemented in 512 * 512 spatial resolution, which is near the officially recommended resolution for backbone and sufficient for demonstration of any method's practicality. Existing DiT blocks are fine-tuned using LoRA. The DG is implemented as a lightweight Vision Transformer (ViT) after comparing several architectures' performance.

\begin{table}[tb]
\centering
\caption{Quantitative comparison with state-of-the-art methods on three face video benchmarks. The best and second-best performances are marked in \textbf{\color{red}red} and \textbf{\color{blue}blue} respectively. Here ours is without DG.}
\label{tab:sota_comparison}
\resizebox{\textwidth}{!}{
\begin{tabular}{@{}c|l|cccccc@{}}
\toprule
Datasets & Metrics &  PGTFormer & SVFR & Vivid-VR & FlashVSR & SeedVR2 & Ours\\ \midrule
\multirow{8}{*}{VFHQ} 
 & PSNR $\uparrow$ & \textbf{\color{blue}25.63} & 25.54 & 19.82 & 19.57 & 19.80 & \textbf{\color{red}26.35}  \\
 & SSIM $\uparrow$ & 0.66 & 0.72 & 0.72 & 0.71 & \textbf{\color{blue}0.73} & \textbf{\color{red}0.74} \\
 & LPIPS $\downarrow$ & 0.30 & 0.22 & 0.21 & \textbf{\color{blue}0.18} & 0.21 & \textbf{\color{red}0.17} \\
 & MUSIQ $\uparrow$ & 64.22 & 65.60 & 71.21 & \textbf{\color{red}75.33} & 60.01 & \textbf{\color{blue}72.54} \\
 & CLIP-IQA $\uparrow$ & \textbf{\color{blue}0.56} & 0.50 & 0.55 & \textbf{\color{red}0.71} & 0.48 & \textbf{\color{blue}0.56} \\
 & IDS $\uparrow$ & 0.79 & \textbf{\color{red}0.87} & 0.83 & 0.86 & 0.85 & \textbf{\color{red}0.87} \\ 
 & LMD $\downarrow$ & 6.055 & 4.71 & 4.64 & 6.10 & \textbf{\color{blue}4.59} & \textbf{\color{red}4.20}  \\ 
 & TLME $\downarrow$ & 5.72 & 4.11 & \textbf{\color{blue}3.98} & 4.29 & 4.19 &  \textbf{\color{red}3.72}  \\ \midrule
\multirow{7}{*}{CelebV-HQ} 
 & PSNR $\uparrow$ & 22.75 & \textbf{\color{blue}24.95} & 23.91 & 24.59 & 24.66 & \textbf{\color{red}25.52}  \\
 & SSIM $\uparrow$ & 0.47 & \textbf{\color{blue}0.71} & 0.64 & 0.68 & 0.70 & \textbf{\color{red}0.75}  \\
 & LPIPS $\downarrow$ & 0.55 & \textbf{\color{blue}0.28} & 0.40 & 0.32 & 0.34 & \textbf{\color{red}0.22} \\
 & MUSIQ $\uparrow$ & 47.1 & 57.73 & 63.81 & \textbf{\color{red}69.69} & 48.72 & \textbf{\color{blue}66.26}  \\
 & CLIP-IQA $\uparrow$ & 0.44 & 0.47 & \textbf{\color{blue}0.54} & \textbf{\color{red}0.70} & 0.44 & \textbf{\color{blue}0.54} \\
 & IDS $\uparrow$ & 0.55 & \textbf{\color{red}0.77} & 0.66 & \textbf{\color{red}0.77} & 0.64 & \textbf{\color{red}0.77}  \\ 
 & LMD $\downarrow$ & 45.05 & \textbf{\color{blue}7.54} & 15.30 & 9.44 & 16.04 & \textbf{\color{red}5.26}  \\ 
 & TLME $\downarrow$ & 29.16 & \textbf{\color{blue}5.34} & 8.59 & 6.34 & 11.39 & \textbf{\color{red}4.01}  \\  \midrule
\multirow{2}{*}{VoxCeleb2} 
 & MUSIQ $\uparrow$ & 58.71 & 63.02 & \textbf{\color{red}73.87} & \textbf{\color{blue}72.57} & 56.62 & 68.41  \\
 & CLIP-IQA $\uparrow$ & 0.49 & 0.52  & \textbf{\color{blue}0.67} & \textbf{\color{red}0.75} & 0.45 & \textbf{\color{blue}0.67}  \\ 
 \midrule
\end{tabular}
}
\end{table}
\subsection{Comparison with State-of-the-Art Methods}
We compare first our method against state-of-the-art (SOTA) FVSR methods including \textbf{PGTFormer}\cite{xu2024beyond} and \textbf{SVFR}\cite{wang2025svfrunifiedframeworkgeneralized}, then against SOTA general VSR methods to more universally prove the performance: \textbf{Vivid-VR}\cite{VividVR}, \textbf{FlashVSR}\cite{FlashVSR} and \textbf{SeedVR2}\cite{wang2025seedvr2}.

\paragraph{\textbf{Performance}}
As shown in Table~\ref{tab:sota_comparison}, our model shows an overall SOTA performance. It achieves the lowest LPIPS on all datasets with GT, with an excellent average lower than 0.2. It has the best performance on all general and facial fidelity metrics, which demonstrates its capabilities in structural consistency, spatio-temporal consistency, and facial feature reconstruction. It remains competitive on no-reference metrics. Given that our approach is based on a small amount of fine-tuning (only 20k iterations), the results undoubtedly validate our conditioning paradigm. Note that FlashVSR attains the best performance on generic perceptual metrics (no-reference), which is reasonable since it has conducted the largest-scale model post-training and auxiliary components' training on a closed-source 160k image-video dataset, generating distribution that most closely resembles real-world data distribution. 

\paragraph{\textbf{Efficiency}}
Among these methods, PGTFormer is originally discriminative, SeedVR2 and Flash
VSR is one-step generative after their large-scale post-training, while SVFR, Vivid-VR and our method are multi-step generative. As shown in Table ~\ref{tab:vfhq_comparison}, discriminative guidance mechanism largely reduces the NFE (Number of Function Evaluations) of our method by dynamically set the starting point, with an average reduction of 76\%. Meanwhile, this framework significantly improved full-reference metrics while slightly declining no-reference metrics, thereby validating our reasoning about that discriminative guidance shifts the perception-distortion trade-off towards fidelity.
\begin{table}[tb]
\centering
\caption{Quantitative result on VFHQ benchmark and the NFE counting. The best and second-best performances are marked in \textbf{\color{red}red} and \textbf{\color{blue}blue}, respectively. $\lambda$ is set to 0.}
\label{tab:vfhq_comparison}
\resizebox{\textwidth}{!}{
\begin{tabular}{@{}l|ccccccccc@{}}
\toprule
Methods & PSNR $\uparrow$ & SSIM $\uparrow$ & LPIPS $\downarrow$ & MUSIQ $\uparrow$ & CLIP-IQA $\uparrow$ & IDS $\uparrow$ & LMD $\downarrow$ & TLME $\downarrow$ & NFE \\
\midrule
SVFR 
& 25.54 & 0.72 & 0.22 & 65.60 & 0.50 & \textbf{\color{red}0.87} & 4.71 & 4.11 & 80 \\
Vivid-VR 
& 19.82 & 0.72 & 0.21 & \textbf{\color{blue}71.21} & \textbf{\color{blue}0.55} & 0.83 & 4.64 & 3.98 & 50 \\
Ours w/o DG 
& \textbf{\color{blue}26.35} & \textbf{\color{blue}0.74} & \textbf{\color{red}0.17} & \textbf{\color{red}71.54} & \textbf{\color{red}0.56} & \textbf{\color{red}0.87} & \textbf{\color{blue}4.20} & \textbf{\color{blue}3.72} & 50 \\
Ours w/ DG
& \textbf{\color{red}26.85} & \textbf{\color{red}0.76} & \textbf{\color{red}0.17} & 65.17 & 0.48 & 0.86 & \textbf{\color{red}4.03} & \textbf{\color{red}3.69} & 12 \\
\bottomrule
\end{tabular}
}
\end{table}

\section{Discussions}
\subsection{Effect of Extracted Face Visual Information}
\begin{figure}[tb]
\begin{minipage}{0.65\linewidth}
        \centering
		\includegraphics[width=0.99\linewidth]{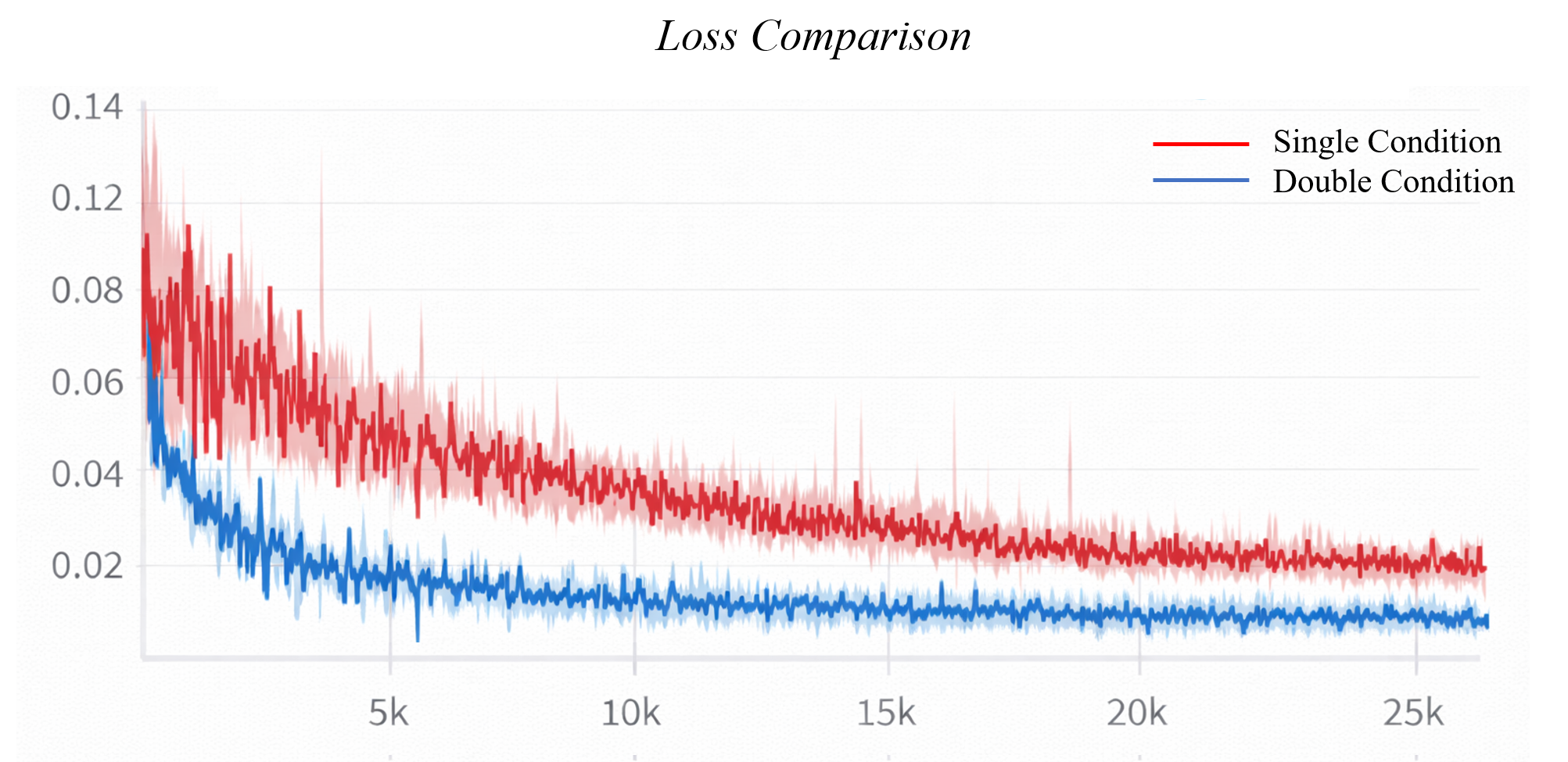}
		\caption{Loss comparison}
		\label{fig5}
\end{minipage}
\begin{minipage}{0.34\linewidth}
\centering
\begin{tabular}{|c|cc|}
\hline
& Single  & Double  \\ \hline
PSNR $\uparrow$ & 22.15 & 26.35   \\ \hline
SSIM $\uparrow$ & 0.61 & 0.74 \\ \hline
LPIPS $\downarrow$ & 0.28 & 0.17 \\ \hline
MUSIQ $\uparrow$ & 69.20 & 71.54 \\ \hline
CLIP-IQA $\uparrow$ & 0.56 & 0.56  \\ \hline
\end{tabular}
\captionof{table}{Performance}
\label{Tab: performance}
\end{minipage}
\end{figure}

We conduct an ablation experiment under identical conditioning paradigm for adapted DiT model. One is trained with the LQ latent as condition (single-condition), the other with also face visual information encoded by DINOv3 (double-condition). Loss evolution (\cref{fig5}) and metrics evaluation (\cref{Tab: performance}) reveal that injection of extracted information condition not only elevates the ultimate performance ceiling but also accelerates model convergence.


\subsection{Discussion on Metrics}
Past works in computer vision have already proved that there is a trade-off between fidelity and perceptual quality: the famous \textbf{perception-distortion} trade-off\cite{p-dtradeoff}. Based on this theory, performance evaluation should use two types of metrics: full-reference metrics for the proximity to GT; no-reference metrics for the alignment with the real-world data distribution and the visual quality.
\begin{figure}[!b]
  \centering
  \begin{subfigure}{0.45\linewidth}
    \includegraphics[width=\linewidth]{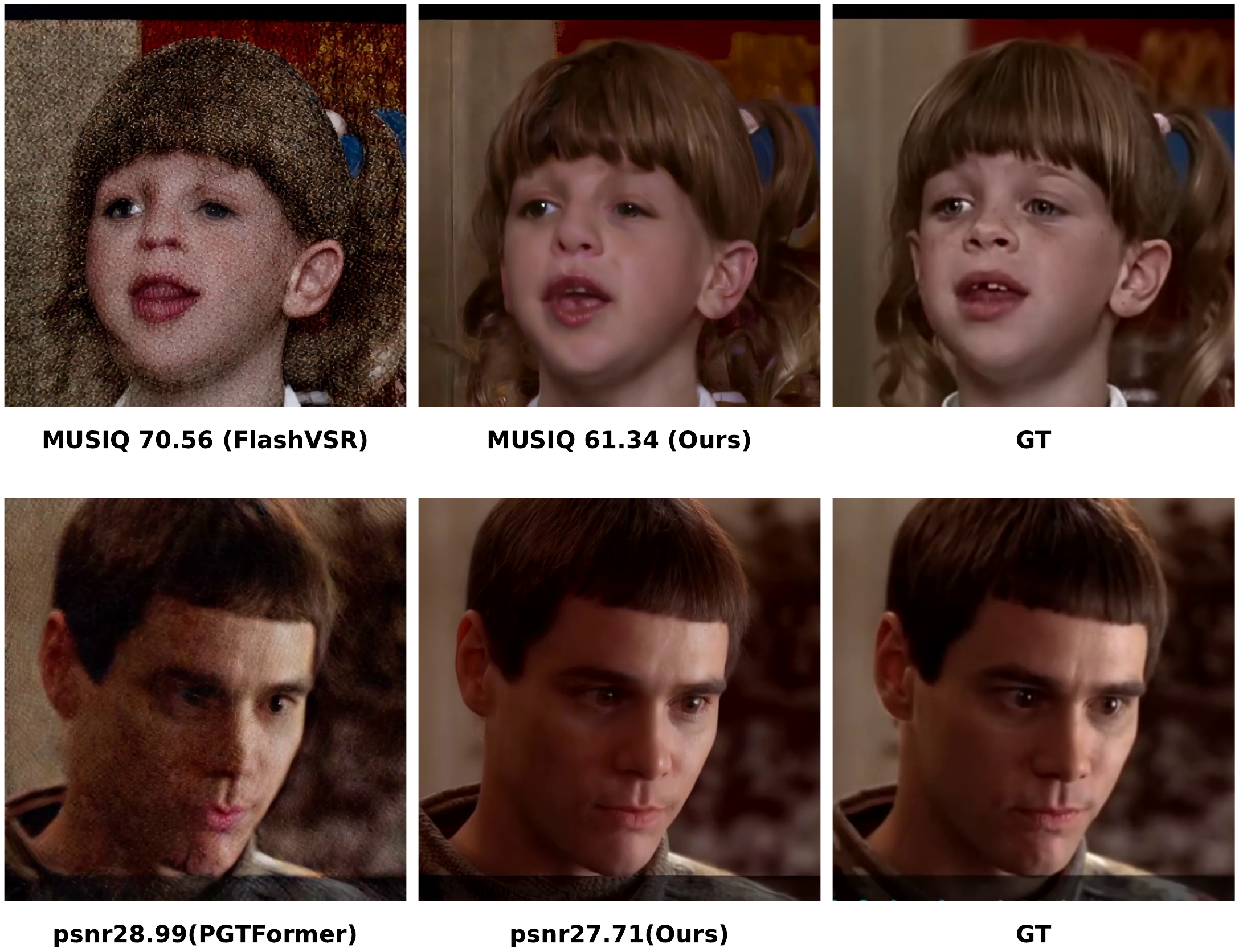}
    \caption{Qualitative Exhibition}
    \label{fig6::a}
  \end{subfigure}
  \hfill
  \begin{subfigure}{0.54\linewidth}
      \includegraphics[width=\linewidth]{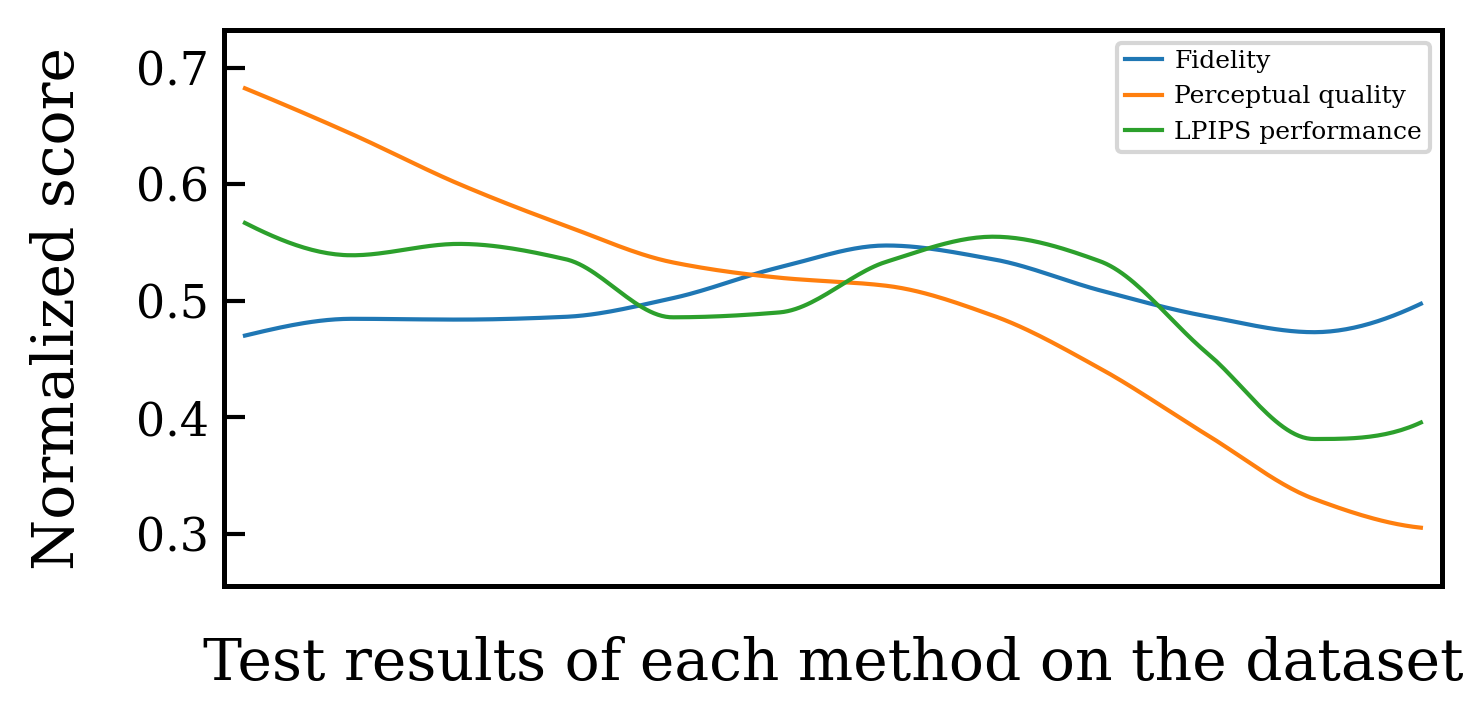}
      \caption{Metrics Relationship}
      \label{fig6::b}
  \end{subfigure}
  \caption{Metrics Discussion. (a): At the \textit{top}, higher MUSIQ sample has evident visual artifacts; at the \textit{bottom}, higher PSNR sample is blurry. (b): The relative normalized score trend among LPIPS, fidelity metrics and perceptual metrics.}
  \label{fig6}
\end{figure}
During evaluation, we identify certain metrics exhibiting unreasonably inflated values in both types, consistent with the trade-off theory: restorations with high no-reference metrics are riddled with repetitive artifacts which is a consequence of overemphasizing realistic details, while restorations with high full-reference metrics appear blurry which reflects mediocrity due to excessive pursuit of global fidelity. Some qualitative examples are shown in \cref{fig6::a}, we confirm that this phenomenon is fairly common across all test data and effectively avoided by our method, which we attribute to the regularization imposed by visual feature condition and discriminative guidance. Among all the metrics, LPIPS demonstrates the most balanced performance in this trade-off and best aligned with human visual perception, because it is only better when both aspects are relatively good (\cref{fig6::b}).

\subsection{Controllable Perception-Distortion Trade-off}
\begin{figure}[tb]
  \centering
  \includegraphics[width=0.6\linewidth]{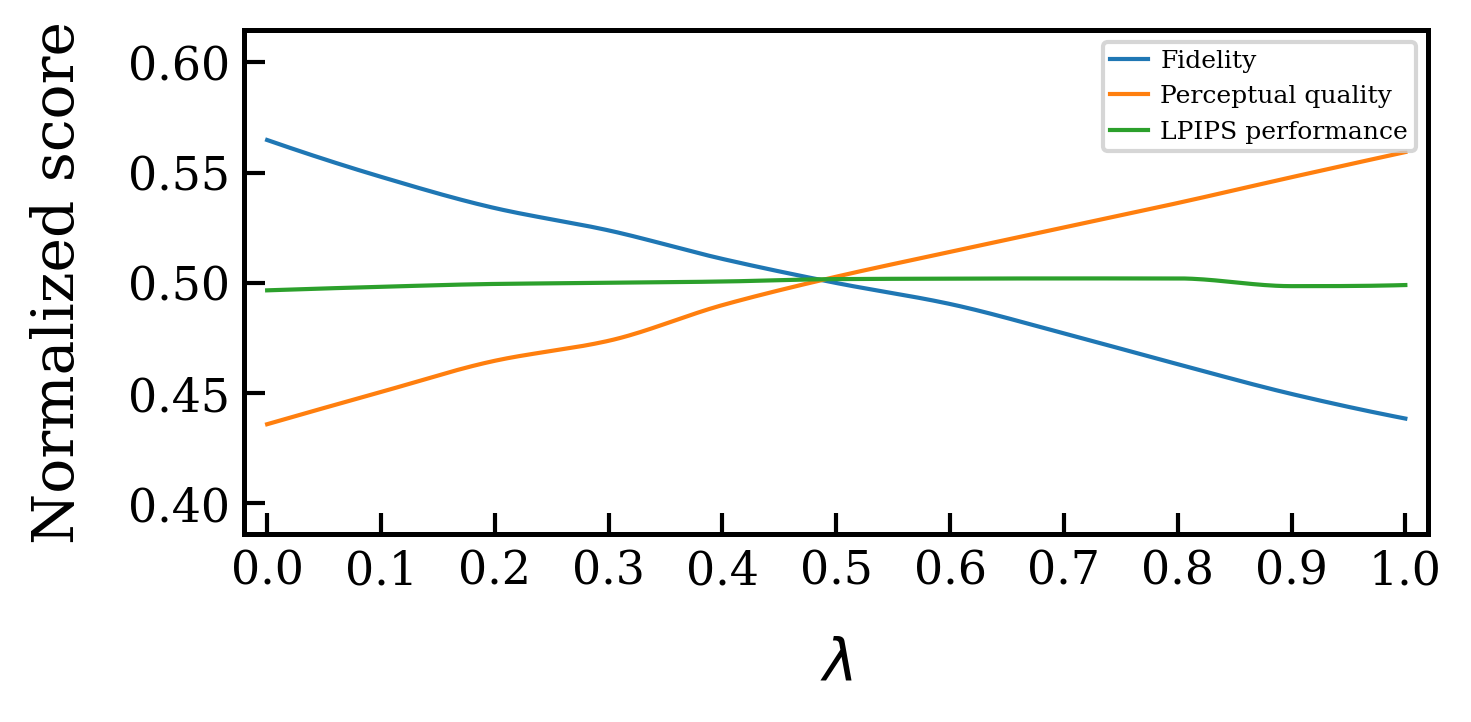}
  \caption{Hyperparameter $\lambda$ for Perception-Distortion Trade-off}
  \label{fig7}
\end{figure}

In \cref{fig7}, normalized scores of metric groups show that the hyperparameter $\lambda$ successfully regulates the perception-distortion trade-off, providing a controllable external interface that can be manipulated as needed, thereby validating our prior deduction about dynamic initialization.
\section{Conclusion}
In this work, we propose DTI, a paradigm allowing dynamic starting for GFVSR, which makes it conform to inherent logic of the restoration task. Our method is built upon a pretrained DiT model with only minor model adaptation and fine-tuning. Dynamic starting is realized by discriminative guidance, with model performance improvement attributed to novel method for condition enhancing and injecting. The proposed DG simultaneously performs coarse-grained low-frequency information anchoring and global noise level prediction, enabling DTI, dynamic NFE reduction and a little shift towards fidelity in perception-distortion trade-off. Extensive experiments on diverse benchmarks demonstrate SOTA performance of our method. The perception-distortion trade-off among metrics is also observed, which illustrates the unreliability of single type of metrics and the comprehensive credibility of LPIPS. 

For DG, one limitation lies in the fact that from-scratch base restricts the generalizability of lightweight training on one dataset. Although we selected the current design after ablation study among LoRA fine-tuning and full-parameter fine-tuning of VAE Encoder, from-scratch training of UNet-like (Encoder-like) NNs, we lack investigation and fine-tuning experiments on pretrained ViT-like models. Leveraging pretrained priors may further improve general performance and training efficiency of DG, which we leave for future work.


\section*{Acknowledgements}
This work is supported by National Natural Science Foundation of China (62271308, 62571322), STCSM (24ZR1432000, 22DZ2229005), 111 plan (BP0719010), and State Key Laboratory of UHD Video and Audio Production and Presentation.
\newpage
%
%
\bibliographystyle{splncs04}
\bibliography{main}
\end{document}